\documentclass[10pt,twocolumn,letterpaper]{article}

\usepackage[pagenumbers]{cvpr}

\usepackage{graphicx,amsmath,amssymb,booktabs,multirow}
\usepackage[pagebackref,breaklinks,colorlinks]{hyperref}

\usepackage[capitalize]{cleveref}
\crefname{section}{Sec.}{Secs.}
\Crefname{section}{Section}{Sections}
\Crefname{table}{Table}{Tables}
\crefname{table}{Tab.}{Tabs.}

\usepackage{xcolor}
\newcommand{\YangWu}[1]{\textcolor{black}{#1}}
\newcommand{\ywu}[1]{\textcolor{black}{#1}}

\def\cvprPaperID{1251}
\def\confName{CVPR}
\def\confYear{2022}

\begin{document}

\title{ACNet: Approaching-and-Centralizing Network for Zero-Shot Sketch-Based Image Retrieval}

\author{Hao Ren\thanks{Both authors contributed equally to this research.} \\
Fudan University\\
Shanghai, China\\
{\tt\small hren17@fudan.edu.cn}
\and
Ziqiang Zheng$^{\ast}$\\
HKUST\\
Hong Kong, China\\
{\tt\small zhengziqiang1@gmail.com}
\and
Yang Wu\\
ARC Lab, Tencent PCG\\
Shenzhen, China\\
{\tt\small dylanywu@tencent.com}
\and
Hong Lu\\
Fudan University\\
Shanghai, China\\
{\tt\small honglu@fudan.edu.cn}
\and
Yang Yang\\
UESTC\\
Chengdu, China\\
{\tt\small dlyyang@gmail.com}
\and
Sai-Kit Yeung\\
HKUST\\
Hong Kong, China\\
{\tt\small saikit@ust.hk}
}

\maketitle
\begin{abstract}
The huge domain gap between sketches and photos and the highly abstract sketch representations pose challenges for sketch-based image retrieval (\underline{SBIR}). The zero-shot sketch-based image retrieval (\underline{ZS-SBIR}) is more generic and practical but poses an even greater challenge because of the additional knowledge gap between the seen and unseen categories. To simultaneously mitigate both gaps, we propose an \textbf{A}pproaching-and-\textbf{C}entralizing \textbf{Net}work (termed ``\textbf{ACNet}'') to jointly optimize sketch-to-photo synthesis and the image retrieval. The retrieval module guides the synthesis module to generate large amounts of diverse photo-like images which gradually approach the photo domain, and thus better serve the retrieval module than ever to learn domain-agnostic representations and category-agnostic common knowledge for generalizing to unseen categories. These diverse images generated with retrieval guidance can effectively alleviate the overfitting problem troubling concrete category-specific training samples with high gradients. We also discover the use of proxy-based NormSoftmax loss is effective in the zero-shot setting because its centralizing effect can stabilize our joint training and promote the generalization ability to unseen categories. Our approach is simple yet effective, which achieves state-of-the-art performance on two widely used ZS-SBIR datasets and surpasses previous methods by a large margin. 
\end{abstract}
 
\section{Introduction}

\begin{figure}
  \centering
  \includegraphics[width=\linewidth]{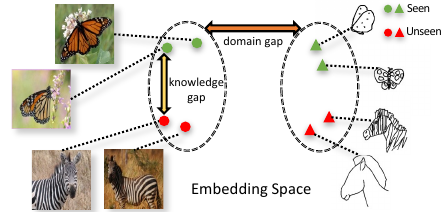}
  \caption{Two challenges in ZS-SBIR: 1) huge domain gap between sketches and photos; and 2) knowledge gap between seen categories and unseen categories.}
  \label{fig:intro}
\end{figure}

Sketch-based image retrieval (SBIR)~\cite{eitz2010sketch, hu2013performance, liu2017deep, wang2019sketch, wang2019enhancing, chaudhuri2020crossatnet} aims to perform cross-domain image retrieval among the photos and sketches drawn \ywu{by} humans. Touch-screen devices (\eg, smartphones and iPad) enable us to draw free-hand sketches conveniently. The drawn sketches are regarded as queries and the retrieval system can resort to returning some relevant photos according to the user’s intent. Considering the lack of colors, textures and detailed structural information, the sketches are highly \textit{iconic}, \textit{succinct} and \textit{abstract}. The huge domain gap and the asymmetrical information between sketches and photos pose \ywu{great} challenges for SBIR. Since category labels provide a shortcut~\cite{li2019zero} for the retrieval model, the retrieval system may overfit the category label hence the retrieval problem would then turn into a classification problem. The follow-up zero-shot sketch-based image retrieval (ZS-SBIR) is introduced in~\cite{yelamarthi2018zero} under a more practical and realistic setting, where the \ywu{test} data are from unseen categories. The knowledge gap between seen categories and unseen categories makes ZS-SBIR more intractable, as shown in \YangWu{Figure}~\ref{fig:intro}. The domain gap and knowledge gap are the two biggest challenges for ZS-SBIR.

To simultaneously mitigate the huge domain gap and \ywu{knowledge gap}, we design a novel, simple and effective \textbf{\ywu{approaching and centralizing}} \ywu{network, which jointly trains sketch-to-photo synthesis and retrieval,} as shown in \YangWu{Figure}~\ref{fig:model}. \ywu{The} sketch-to-photo synthesis \ywu{module encourages} the retrieval module to \ywu{focus more on domain-agnostic} information \ywu{for proper similarity measurement. This is done by constantly refining and feeding the} synthesized photo-like images \ywu{into} the retrieval module during the training phase. Even \ywu{though there are some} noise and uncertainty introduced along with the synthesis, \ywu{the continuously generated and refined images are of} high diversity\ywu{, which gradually \textbf{approach} the photo domain and thus benefit training a robust retrieval module that can overcome the domain gap}.

\ywu{The large amounts of synthesised images with} high diversity \ywu{can also} force the retrieval module to better learn the common knowledge amongst the seen and unseen categories. \ywu{It prevents} the retrieval module \ywu{from extracting} the category-specific information and \ywu{overfitting/remembering} the \ywu{specific details} of the concrete training samples, which is catastrophic under the zero-shot setting. \ywu{Therefore, our framework can also mitigate the knowledge gap and make the trained model generalize better.}

Furthermore, we \ywu{utilize the} NormSoftmax~\cite{zhai2019classification} loss \ywu{which has a class \textbf{centralization} effect} to stabilize our joint training and promote the generalization capability to unseen categories. Previous pairwise losses (\eg, Contrastive~\cite{hadsell2006dimensionality} and Triplet~\cite{ge2018deep}) aim to optimize the model by sampling the informative pairs or triplets in one batch. It is easy for these methods to overfit the category-specific feature representations when the sampled training pairs or triplets have high gradients. The NormSoftmax assigns one proxy as the center point for each category and optimizes the model with the gradients from all the samples belonging to the assigned proxy. This loss function can better coordinate the relationship between all the samples from each category. Additionally, the noise and uncertainty introduced with the synthesis may cause difficulty in optimizing the pairwise losses. In contrast, the adopted NormSoftmax loss can better alleviate the influence of noise and uncertainty by proxy-based optimization~\cite{movshovitz2017no, teh2020proxynca++}. The \ywu{proposed approaching (via} sketch-to-photo synthesis\ywu{)} and \ywu{centralizing (with} the NormSoftmax loss\ywu{) are} mutually-beneficial, achieving a new state-of-the-art retrieval performance on two widely used ZS-SBIR datasets. To facilitate the development of ZS-SBIR, our codes will be released with the publication of this paper. Extensive ablation studies have been conducted to dissect our method. 

Our main contributions can be summarized as follows:
\begin{itemize}
  \item We propose an ``Approaching-and-Centralizing Network (ACNet)'' to integrate sketch-to-photo synthesis and \ywu{retrieval} through a joint training manner, in which the synthesis module \ywu{can} better serve the downstream retrieval module \ywu{than ever, and the retrieval module is able to guide the synthesis module to generate refined images gradually approaching the photo domain (mitigating the domain gap)}. 
  \item We adopt the NormSoftmax loss to stabilize our joint training and promote the generalization ability under the zero-shot setting\ywu{, thanks to its centralizing effect driven by} proxy-based optimization.
  \item Comprehensive ZS-SBIR experiments and ablation studies on Sketchy Extended~\cite{yelamarthi2018zero} and TU-Berlin Extended~\cite{shen2018zero} datasets \ywu{demonstrating} the superiority of the proposed \ywu{ACNet}.
\end{itemize}

\section{Related Work}
\subsection{SBIR}
SBIR~\cite{eitz2010sketch} has been studied for decades due to its commercial and realistic applications~\cite{chen2009sketch2photo}. Attempts for solving the SBIR task mostly focus on bridging the domain gap between the sketches and photos, which can roughly be grouped in hand-crafted~\cite{eitz2010sketch, hu2013performance} and cross-domain deep learning-based methods~\cite{hadsell2006dimensionality, hermans2017defense, ge2018deep, wang2019multi, torres2021compact, fuentes2021sketch}. Hand-crafted methods~\cite{eitz2010sketch, hu2013performance} mostly work by extracting the edge map from natural photo images and then matching them with sketches using a Bag-of-Words model. Due to the large success and the robust recognition performance of deep learning methods, various specifically designed neural networks~\cite{liu2017deep, lu2021domain, wang2019enhancing, wang2019sketch, chaudhuri2020crossatnet, sain2021stylemeup} have been proposed for SBIR. The classical ranking losses, like contrastive loss~\cite{hadsell2006dimensionality, wang2019multi}, triplet loss~\cite{hermans2017defense, ge2018deep} or classification loss~\cite{zhai2019classification} have also been introduced. 
\begin{figure*}
  \centering
  \includegraphics[width=0.9\linewidth]{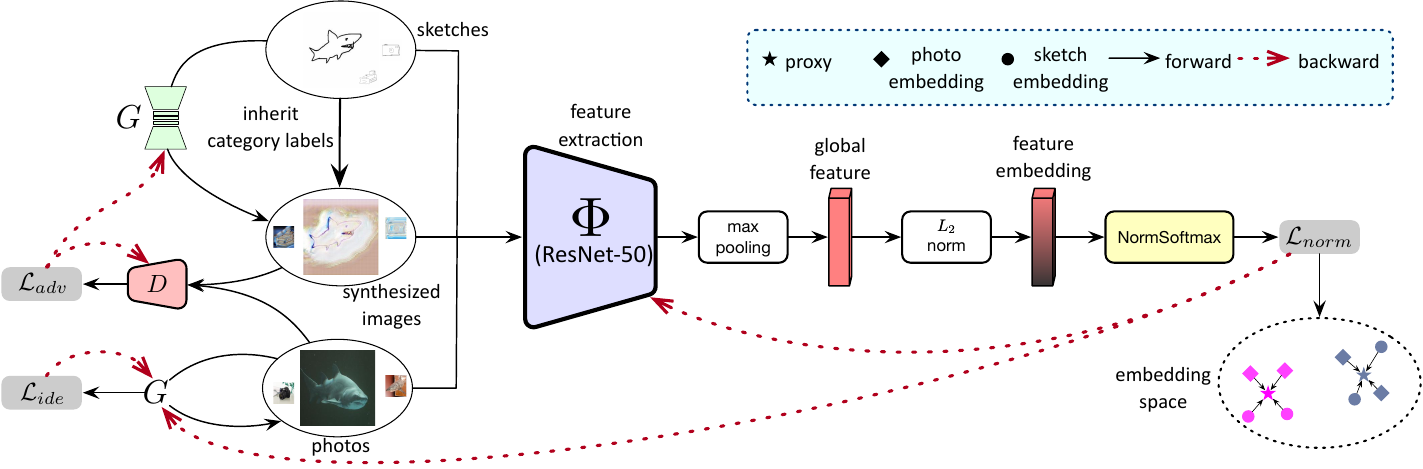}
  \caption{Overview of the proposed \ywu{ACNet}. The sketch-to-photo generator $G$ aims to translate the sketches to the photo-like images while $D$ is to distinguish whether the generated images are real photos and compute $\mathcal{L}_{adv}$. The identity loss $\mathcal{L}_{ide}$ helps generate more photo-like images by forcing $G$ to reconstruct the real photos. The main pipeline serves as the standard process for embedding learning. The embedding from the sketches, synthesized images and photos of the same class are enforced to be close to the assigned proxy, and far away from other proxies under the constraint of $\mathcal{L}_{norm}$. Different colors indicate the different classes. We design a joint training manner to integrate $G$ and $\Phi$. We refer the readers to check the forward and backward procedures to better understand our joint training scheme.}
  \label{fig:model}
\end{figure*}

\subsection{ZS-SBIR}
SBIR requires all test categories to be seen during training, which cannot be guaranteed in real-world applications. The more challenging, generic and practical ZS-SBIR~\cite{shen2018zero} task has attracted the attention of the computer vision community due to its real-world applications, in which the test categories do not appear in the training stage. Recent research~\cite{yelamarthi2018zero, dutta2019semantically, dutta2019style, deng2020progressive, zhang2020zero, tursun2021efficient, wang2021norm} are exploring solutions for projecting the sketches and photos into a shared semantic space to perform accurate cross-domain image retrieval. However, the huge domain gap and the highly abstract sketch representations make it very difficult to perform ideal feature-level content-style disentanglement~\cite{dutta2019semantically, li2019zero, dutta2019style, deng2020progressive} or bi-direction synthesis~\cite{liu2020unsupervised}. To bridge the knowledge gap between seen and unseen categories, existing methods~\cite{dey2019doodle, liu2019semantic, zhang2020zero, zhu2020ocean, deng2020progressive, xu2021progressive} introduced the semantic embeddings from the extra annotations to generate class prototypes for presenting the relationships between semantic categories in the common space. However, there is no explicit connection between the feature representations extracted from visual images and the semantic embeddings borrowed from the extra annotations. Furthermore, the semantic embeddings from the language models~\cite{goldberg2014word2vec, pennington2014glove} are computed based on word co-occurrence frequency. Sometimes these embeddings are not reliable~\cite{wang2021norm} and cannot express the intra-class visual variation. In contrast to previous methods, we introduce the one-directional sketch-to-photo synthesis to mitigate the domain gap and promote the generalization ability in a joint training manner.

\subsection{Sketch-to-photo Synthesis}
Sketch-to-photo synthesis~\cite{eitz2011photosketcher, liu2017unsupervised, zhu2017unpaired, chen2018sketchygan, liu2020unsupervised, wang2021sketch} is a notably challenging task in the field of computer vision, which aims to generate photo-realistic images from the given abstract and exaggerated sketches. Sketch2Photo~\cite{chen2009sketch2photo} proposed to compose new photo images through fusing the retrieved photo images from the given sketch. The semantic segmentation~\cite{arbelaez2012semantic} and image blending~\cite{gracias2009fast} techniques were introduced to achieve the photo editing according to the user's goal. The first deep learning-based free-hand sketch-to-photo synthesis is SketchyGAN~\cite{chen2018sketchygan}, which aims to optimize an encoder-decoder model based on the aligned sketch-photo pairs. Ghosh \etal~\cite{ghosh2019interactive} proposed a multi-class photo generation based on incomplete edges or sketches. This method adopted a two-stage strategy for shape completion and appearance synthesis, which highly relies on paired training data. Sketchformer~\cite{ribeiro2020sketchformer} designed a sequential sketch-to-photo generation model to promote the naturalness of the photo images. Liu \etal~\cite{liu2020unsupervised} conducted unsupervised sketch-to-photo synthesis and further analyzed the potential of adopting the synthesized images for retrieval. \ywu{Ideally, a perfect sketch-to-photo generator could synthesize the desired photo images with distinguishable and reliable feature representations for more accurate image retrieval, whilst preserving the intra-class and inter-class distribution after translation. However, the unsupervised domain translation performance is plagued by the huge domain gap as well as the highly abstract sketch representations. The translated photo images still suffer from visual artifacts~\cite{li2021learning} and noise even utilizing the semantic priors~\cite{liu2017deep, zhang2018generative, li2021learning} and laborious generative networks~\cite{ribeiro2020sketchformer, zheng2020forkgan, li2021learning}. Since the two modules are optimized separately, the noise and uncertainty introduced in the synthesis module could be propagated to the retrieval module and the error accumulation could heavily restrict the retrieval performance. Even if the generated images look realistic to humans, their benefits may not be able to surpass harms to the downstream retrieval task. Differently, our proposed ACNet jointly optimizes synthesis and retrieval, and thus ensures a significant performance boost.}

\section{Method}
\subsection{Problem Formulation}
Consider $n$ photos and $m$ sketches denoted as $\mathcal{P}=\{(p_i,y_{p_i})|y_{p_i}\in\mathcal{Y}\}_{i=1}^{n}$, and $\mathcal{S}=\{(s_i,y_{s_i})|y_{s_i}\in\mathcal{Y}\}_{i=1}^{m}$ respectively. Under the SBIR setting, $\mathcal{S}$ and $\mathcal{P}$ are divided into the training set and \ywu{test} set with same category label set $\mathcal{Y}$. SBIR is to retrieve the best matched $p_j\in \mathcal{P}$ based on a query sketch $s_i$ in $\mathcal{S}$, such that $y_{s_i} = y_{p_j}$. Under the ZS-SBIR setting, $\mathcal{Y}$ is split into $\mathcal{Y}_{tra}$ and $\mathcal{Y}_{test}$, in which there is no category overlap between $\mathcal{Y}_{tra}$ and $\mathcal{Y}_{test}$ ($\mathcal{Y}_{tra}\cap \mathcal{Y}_{test}=\emptyset$). The training data are $\mathcal{S}_{tra}=\{(s_i,y_{s_i})|y_{s_i}\in \mathcal{Y}_{tra}\}$, $\mathcal{P}_{tra}=\{(p_i,y_{p_i})|y_{p_i}\in \mathcal{Y}_{tra}\}$, and the \ywu{test} data are $\mathcal{S}_{test}=\{(s_i,y_{s_i})|y_{s_i}\in \mathcal{Y}_{test}\}$, $\mathcal{P}_{test}=\{(p_i,y_{p_i})|y_{p_i}\in \mathcal{Y}_{test}\}$. The ZS-SBIR model is trained on data $(\mathcal{S}_{tra},\mathcal{P}_{tra})$, and tested on $(\mathcal{S}_{test},\mathcal{P}_{test})$.

\subsection{Main Pipeline}
We refer the readers to check the overall architecture of the proposed method in \YangWu{Figure~\ref{fig:model}}. 

\noindent\textbf{\ywu{Approaching by} Sketch-to-photo Synthesis}. Suppose the sketch $s_i$ from $\mathcal{S}_{tra}$ and the photo $p_j$ from $\mathcal{P}_{tra}$, we first aim to generate a photo-like image $s_i^{*}=G(s_i)$ based on $s_i$ through a generator $G:\mathcal{S}_{tra}\rightarrow\mathcal{P}_{tra}$. The adversarial loss of the GAN architecture can be expressed as:
\begin{equation}
\begin{split}
\mathcal{L}_{adv}(G,D)=&\mathbb{E}_{s_i,p_j\sim P_{data}(\mathcal{S}_{tra},\mathcal{P}_{tra})}\left[\log D(p_j)\right]+ \\
& \mathbb{E}_{s_i\sim P_{data}(\mathcal{S}_{tra})}\left[\log (1-D(G(s_i)))\right],
\end{split} 
\label{eq:gan}
\end{equation}
where $D$ is the discriminator to distinguish whether the image is from the real photo domain. The goal is to learn a mapping function, which could generate photo-like images that match the real photo distribution $P_{data}(\mathcal{P}_{tra})$. After the sketch-to-photo synthesis, we assign the category label of $s_i$ to $s_i^{*}$. It is non-trivial to define such label-preserving synthesis-based transformations, especially when the uncertainty and noise have been introduced with the image synthesis. The synthesized images possess more texture and RGB information and thus \ywu{gradually approach} the photo domain, which can better serve cross-domain image retrieval. Considering there is no pixel-level constraint for $G$, $G$ would tend to generate images with visual artifacts. To alleviate this problem, we design an identity loss $\mathcal{L}_{ide}$ between $p_j$ and $G(p_j)$, which is expressed as:
\begin{equation}
    \mathcal{L}_{ide}=\mathbb{E}_{\mathcal{P}_{tra}}\|G(p_j)-p_j\|_{1}.
\end{equation}
Since $s_i$ and $p_j$ share similar semantic contents (\eg, the category and structure information), we can boost the synthesis performance by reconstructing the photos at the same time. With the full supervision from $\mathcal{L}_{ide}$, we could generate more photo-like images.

\noindent\textbf{Feature Extraction}. The sketch, generated image and real photo are fed into the backbone network to extract feature representations. Like previous methods, we adopt ResNet-50~\cite{he2016deep} (denoted as $\Phi$) as backbone. The outputs after max pooling are transformed into the desirable embedding dimension through a fully connected layer. $L_2$\ywu{-norm} is adopted to obtain the final embedding for the retrieval task.

\noindent\textbf{\ywu{Centralizing with NormSoftmax}}. The NormSoftmax loss~\cite{zhai2019classification} is used as our objective function to increase the inter-class distance and reduce the \YangWu{intra}-class distance over the photo and sketch set. Each category is assigned a learnable proxy. The final embedding $x$ is enforced to be close to the proxy of its category, and far away from other proxies, as shown in \YangWu{Figure}~\ref{fig:model}. The objective function is:
\begin{equation}
  \mathcal{L}_{norm}(x) = -log(\frac{exp(\frac{x^Tp_y}{t})}{\sum_{z \in Z}exp(\frac{x^Tp_z}{t})}),
\end{equation}
where $Z$ is the set of all proxies, $p_y$ is the target proxy, $t$ is temperature scale. We set $t=0.05$ following the default setting in~\cite{zhai2019classification}. By assigning the label of $s_i$ to $s_i^{*}$, we design \ywu{a} chainer loss described as:
\begin{equation}
    \mathcal{L}_{cha}=\mathcal{L}_{norm}(\Phi(s_i^{*})),
\end{equation}
\ywu{with which} we can better reduce the domain gap between the sketch and photo domain through the intermediate synthesized images. Besides, we also compute the NormSoftmax loss for the real sketches and photos. The final NormSoftmax loss is computed as:
\begin{equation}
    \mathcal{L}_{norm}=\mathcal{L}_{cha}+\mathcal{L}_{norm}(\Phi(s_i))+\mathcal{L}_{norm}(\Phi(p_j)).
    \label{eq:norm}
\end{equation}

\noindent\textbf{Final Objective Function}. The final objective function for $G$, $D$ and $\Phi$ is described as:
\begin{equation}
    \mathcal{L}(G,D,\Phi)=\mathcal{L}_{adv}(G,D)+ \lambda\mathcal{L}_{norm}+\gamma\mathcal{L}_{ide},
    \label{eq:all}
\end{equation}
where $\lambda$ and $\gamma$ are hyper-parameters to balance the contribution of each component. We set $\lambda=10$ and $\gamma=0.1$ in our experiments and provide comprehensive experiments using different values of $\lambda$ and $\gamma$ in appendix. 

\noindent\textbf{\ywu{Joint Approaching and Centralizing}}. We optimize $G$ and $\Phi$ through a \ywu{\textbf{joint training}} manner and the synthesized images are constantly fed into $\Phi$. Through the sketch-to-photo synthesis, we could generate sufficient photo-like examples with high data diversity and force $\Phi$ to extract more reliable and effective feature representations under the constraint of $\mathcal{L}_{cha}$. Besides, by sufficiently generating samples in the latent space, we could promote the generalization ability to unseen categories without requiring access to data from those categories. Our framework is \textbf{model-agnostic} and could choose various GAN architectures for synthesis and backbone networks for feature extraction.

\noindent\textbf{Inference}. In the test phase, $G$ first generates one photo-like image based on the sketch query and the generated image is used for the retrieval among the photos. We do \textbf{not} conduct reverse photo-to-sketch synthesis since we are performing sketch-based image retrieval and the photos are not available in the inference time. 

\section{Experiments}
\subsection{Experimental Setup}
\noindent\textbf{Datasets}. \textbf{Sketchy Extended} dataset contains 75,481 sketches and 73,002 photos (12,500 images from~\cite{sangkloy2016sketchy} and 60,502 images from ImageNet~\cite{deng2009imagenet} organized by Liu \etal~\cite{liu2017deep}) from 125 categories. We follow the same zero-shot data partitioning as~\cite{yelamarthi2018zero}, in which 21 unseen classes from ImageNet for testing and other classes for training. \textbf{TU-Berlin Extended}~\cite{eitz2012humans} contains 20,000 sketches evenly distributed over 250 object categories. 204,070 photo images collected by Liu \etal~\cite{liu2017deep} are included. The partition protocol introduced in~\cite{shen2018zero} is used for creating zero-shot training and \ywu{test} sets. 30 randomly picked classes that include at least 400 photo images are used for testing, and other classes are used for training. 
\begin{table}
\centering
\caption{The ZS-SBIR performance comparison under various settings on Sketchy Extended dataset: 1) Triplet and NormSoftmax for optimization; and 2)  with or without sketch-to-photo synthesis through CycleGAN to mitigate the domain gap.}
\label{tab:prior_sketchy}
\begin{tabular}{c|c|cccc}
\toprule
\begin{tabular}[c]{@{}c@{}}Loss\\ Type\end{tabular} & Synthesis & \begin{tabular}[c]{@{}c@{}}mAP\\ @200\end{tabular} & \begin{tabular}[c]{@{}c@{}}mAP\\ @all\end{tabular} & \begin{tabular}[c]{@{}c@{}}Prec\\ @100\end{tabular} & \begin{tabular}[c]{@{}c@{}}Prec\\ @200\end{tabular} \\ \midrule
\multirow{2}{*}{$\mathcal{L}_{triplet}$} & -   & \textbf{35.5}     & \textbf{40.8}   & \textbf{51.2}       &  \textbf{47.3}         \\
 & \checkmark   & 33.1     & 38.2       & 48.1       & 44.4         \\\midrule
\multirow{2}{*}{$\mathcal{L}_{norm}$} & -  & \textbf{45.2}  & \textbf{48.6}       & \textbf{60.2}       & \textbf{55.7}         \\
 & \checkmark & 43.5     & 47.5       & 58.0     & 53.9         \\ \bottomrule
\end{tabular}
\end{table}

\noindent\textbf{Implementation Details}. Following the previous methods~\cite{yelamarthi2018zero, dutta2019semantically, dey2019doodle, dutta2019style, liu2019semantic, zhang2020zero, wang2021norm}, ResNet-50~\cite{he2016deep} are adopted as the backbone network. The image resolution and the batch size are $224\times 224$ and 64, respectively. For $G$, we choose the same architecture of CycleGAN~\cite{zhu2017unpaired}. Please note that we only design the forward sketch-to-photo generator. Furthermore, to reduce the inference time and computational cost, we modify the channel number of the first convolutional layer to 8, and the number of residual blocks to 8, so that our generator is very lightweight. We adopt PatchGAN discriminator~\cite{isola2017image} architecture for $D$ and also set the channel number of the first convolutional layer to 8 to alleviate the memory burden and computational costs. We provide the detailed neural network configurations in appendix. All the models are warm-up with 1 epoch and have been optimized with 10 epochs. We choose Adam~\cite{kingma2015adam} with learning rate $1e^{-3}$ for optimization. Our code is implemented with PyTorch~\cite{paszke2019pytorch} library and the experiments are conducted on the Geforce RTX 3090 GPU. 

\noindent\textbf{Evaluation Metrics}. Precision ($Prec$) and mean Average Precision ($mAP$) are two main metrics for the evaluation of ZS-SBIR task~\cite{shen2018zero}. $Prec$ is calculated for top $k$ (\ie, 100, 200) ranked results, and $mAP$ is computed for top $k$ or all ranked results. Higher $Prec$ and $mAP$ indicate better retrieval performance.

\begin{figure}
  \centering
  \includegraphics[width=0.9\linewidth]{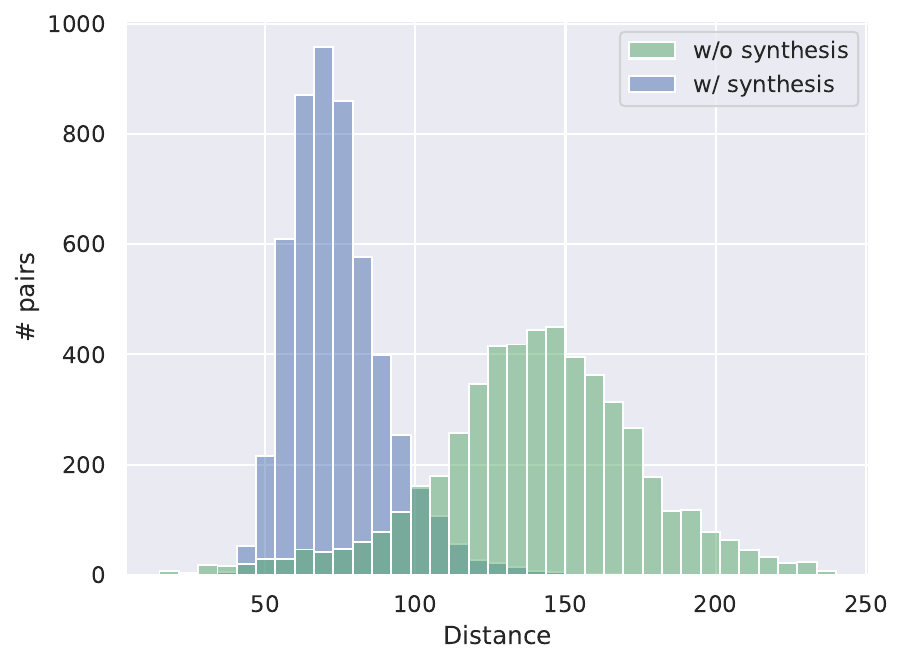}
  \caption{Histograms of sample distances under two settings: 1) between original sketches and real photos (``w/o synthesis'') and 2) between synthesized images and real photos (``w/ synthesis'').}
  \label{fig:prior_sketchy}
\end{figure}

\begin{figure*}
  \centering
  \includegraphics[width=\linewidth]{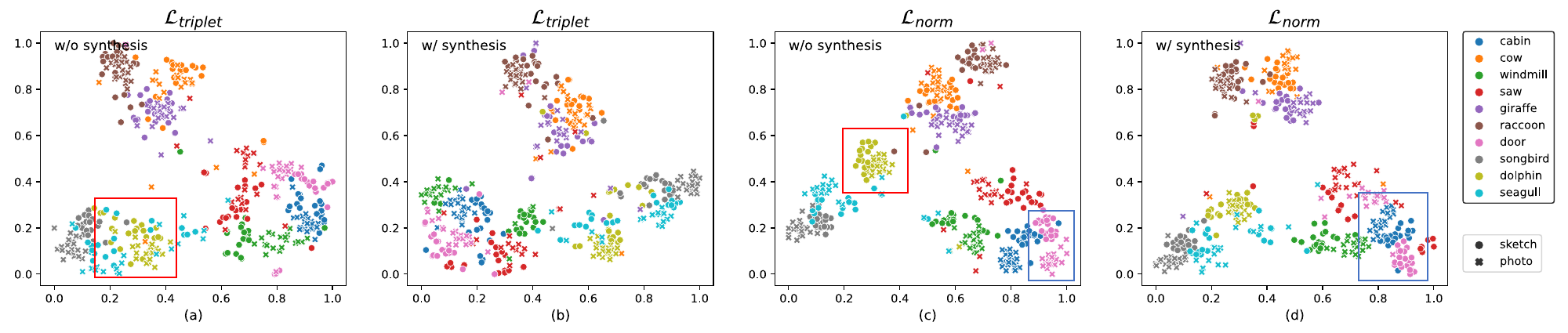}
  \caption{T-SNE visualization of sketch and photo embeddings on Sketchy Extended dataset. We randomly choose 20 samples from 10 unseen test categories for visualization. Different colors refer to different categories. The two-stage training strategy cannot obtain more separable embeddings than the vanilla setting, regardless of using Triplet loss or NormSoftmax loss. We refer the readers to pay more attention on the regions covered by same color boxes for better comparison.}
  \label{fig:tsne_sketchy}
\end{figure*}

\begin{table*}
\centering
\caption{Overall comparison of our method and other approaches on Sketchy Extended and TU-Berlin Extended datasets. ``$\dagger$''denotes results obtained by hashing codes, and ``-'' means that corresponding results are not reported in the original papers. The best and second-best results are bold and underlined, respectively.}
\label{tab:results}
\begin{tabular}{c|c|c|c|cc|cc}
\toprule
\multirow{2}{*}{Methods} & \multirow{2}{*}{Venue} & \multirow{2}{*}{Semantic} &
\multirow{2}{*}{Dim} & \multicolumn{2}{c|}{Sketchy Extended} & \multicolumn{2}{c}{TU-Berlin Extended} \\
       &        &         &     & mAP@200 & Prec@200 & mAP@all & Prec@100 \\ \midrule
DSH{\cite{liu2017deep}}      & CVPR'17   & $\times$  & $64^\dagger$     & 5.9     & 15.3   & 12.2       & 19.8            \\
CVAE{\cite{yelamarthi2018zero}}      & ECCV'18   & $\times$  & 4096     & 22.5     & 33.3   & 0.5       & 0.1            \\
CSDB{\cite{dutta2019style}}        & BMVC'19   &  $\times$   & 4096    & 35.8     & 40.0    & 25.4       & 35.5           \\
DSN{\cite{wang2021domain}}   & IJCAI'21       &    $\times$  & 512     & -     & -           & 48.1     & 58.6      \\ 
NAVE{\cite{wang2021norm}}   & IJCAI'21       &     $\times$     & 512   & -     & -     & 49.3      & 60.7           \\ 
\multirow{2}{*}{RPKD{\cite{tian2021relationship}}}   & \multirow{2}{*}{ACM MM'21}   &  \multirow{2}{*}{$\times$}     & $64^\dagger$   & 37.1     & 48.5     & 36.1      & 49.1    \\
&  &      & 512   & \underline{50.2}     & \underline{59.8}     & 48.6      & 61.2    \\ \midrule
ZSIH{\cite{shen2018zero}}   & CVPR'18  & \checkmark  & $64^\dagger$    & -      & -       & 22.0     & 29.1    \\
\multirow{2}{*}{SEM-PCYC{\cite{dutta2019semantically}}}  & \multirow{2}{*}{CVPR'19}  & \multirow{2}{*}{\checkmark}   & $64^\dagger$       & -        & -      & 29.3      & 39.2     \\
&  &  & 64      & 47.0        & 43.7      & 29.7      & 42.6     \\
Doodle{\cite{dey2019doodle}}       & CVPR'19        & \checkmark &   4096          & 46.1      & 37.0    & 10.9        & -      \\
\multirow{2}{*}{SAKE{\cite{liu2019semantic}}}     & \multirow{2}{*}{ICCV'19}       &   \multirow{2}{*}{\checkmark}    & $64^\dagger$      & 35.6      & 47.7        & 35.9      & 48.1    \\ 
&      &   & 512      & 49.7      & \underline{59.8}        & 47.5      & 59.9    \\ 
OCEAN{\cite{zhu2020ocean}}         & ICME'20   &  \checkmark & 512       & -      &   -      & 33.3      & 46.7     \\
PDFD{\cite{xu2021progressive}}  & IJCAI'20     &    \checkmark   & 512    & -     & -  & 48.3     & 60.0   \\ \midrule
\multirow{2}{*}{\textbf{\ywu{ACNet (Ours)}}}     & \multirow{2}{*}{-}       &   \multirow{2}{*}{$\times$}    & $64^\dagger$      &  48.2    &      58.1   &   \underline{53.3}   & \underline{63.8}   \\ 
&     &  & 512      &   \textbf{51.7}  &  \textbf{60.8}     & \textbf{57.7}      & \textbf{65.8}    \\ \bottomrule
\end{tabular}
\end{table*}

\subsection{Triplet vs. NormSoftmax}
We target to demonstrate that the NormSoftmax loss~\cite{zhai2019classification} is more effective than the Triplet loss~\cite{ge2018deep} in the zero-shot setting. We conduct the ZS-SBIR experiments on the Sketchy Extended dataset and the quantitative results are reported in Table~\ref{tab:prior_sketchy}. To make a fair comparison, all the hyper-parameters are set to the same. The Triplet loss can only achieve 35.5\% $mAP@200$ while the NormSoftmax loss has achieved 45.2\% $mAP@200$, which has gained a large performance improvement. The proxy-based optimization framework could centralize all the samples that belong to the same proxy and prevent the model from only remembering some category-specific samples that have high gradients, which is catastrophic for the generalization ability to unseen categories.

\subsection{Two-stage Training vs. Joint Training}
In this section, we aim to demonstrate the limitation of the previous two-stage training, which is adopted to mitigate the domain gap between the sketches and photos for the ZS-SBIR. For the first synthesis part, we adopt the vanilla CycleGAN~\cite{zhu2017unpaired} to perform the unpaired image-to-image (I2I) translation between the sketches and photos\footnote{We adopt the official implementation \url{https://github.com/junyanz/pytorch-CycleGAN-and-pix2pix} following the default setting.}. The train/test split strictly follows the ZS-SBIR setting. After the unpaired I2I model converges, we adopt the trained sketch-to-photo generator for inference and translate all the sketch images into photo-like images. Please note, both training and \ywu{test} sketch images have been translated to the photo domain for further training and testing. We then perform the ZS-SBIR experiments based on the synthesized photo-like images and the real photo images. The original category labels of the sketches are inherited for training. We choose both Triplet loss and the adopted NormSoftmax loss for optimization and the quantitative results are also reported in Table~\ref{tab:prior_sketchy}. Compared with the counterpart results achieved on the vanilla setting (between the original sketches and photos), there is a slight performance drop regardless of using Triplet loss or NormSoftmax loss. 

\begin{figure}
  \centering
  \includegraphics[width=\linewidth]{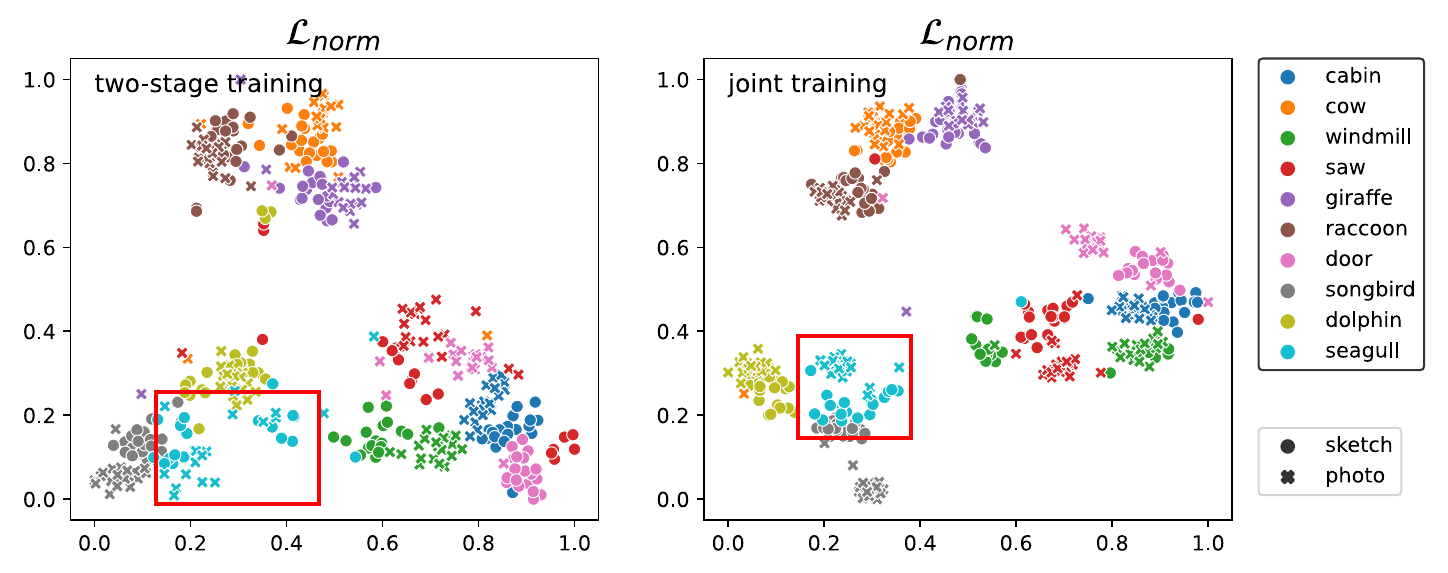}
  \caption{T-SNE visualization of sketch and photo embeddings on Sketchy Extended dataset under the two-stage training and joint training. We refer the readers to pay more attention on the regions covered by same color boxes for better comparison.}
  \label{fig:sketchy_ours}
\end{figure}

To dissect this failure, we compute the $L_1$ distances between 50 randomly sampled instances from each training category under two settings: 1) between the original sketches and the real photos and 2) between the synthesized images and the real photos. We provide the distance histogram in \YangWu{Figure}~\ref{fig:prior_sketchy} to show the domain distance. With the synthesis, the distances have been reduced a lot, which demonstrates that the sketch-to-photo synthesis has indeed effectively mitigated the domain gap. Furthermore, we provide the T-SNE visualization of sketch and photo embeddings on Sketchy Extended dataset in \YangWu{Figure}~\ref{fig:tsne_sketchy} under the four settings of Table~\ref{tab:prior_sketchy}. The two-stage training strategy cannot obtain more separable embeddings to achieve better ZS-SBIR performance even the synthesis can heavily reduce the domain gap. We attribute this failure to the reason that the gradients of the retrieval module cannot be directly been utilized for the optimization of the synthesis module. Thus, the synthesis module has no idea how to generate images, which can better serve the image retrieval. We provide a direct comparison between the distribution of the embeddings from the two-stage training and the proposed joint training in \YangWu{Figure}~\ref{fig:sketchy_ours}. As illustrated, our joint training has a strong ability to split the embeddings from different classes, which leads to better ZS-SBIR performance.

\begin{table}
\setlength\tabcolsep{3pt}
\centering
\caption{Ablation studies for the proposed method on TU-Berlin Extended dataset. We adopt the ResNet-50 as our backbone and the embedding dimension is 512. The NormSoftmax loss is used as a baseline.}
\label{tab:ablation_tuberlin}
\scalebox{0.8}{\begin{tabular}{cccc|cccc}
\toprule
$\mathcal{L}_{norm}$  & $\mathcal{L}_{adv}\&\mathcal{L}_{cha}$   & $\mathcal{L}_{ide}$  & $\mathcal{L}_{norm}(\Phi(s_i))$  & \begin{tabular}[c]{@{}c@{}}mAP\\ @200\end{tabular}   &  \begin{tabular}[c]{@{}c@{}}mAP\\ @all\end{tabular}   & \begin{tabular}[c]{@{}c@{}}Prec\\ @100\end{tabular}  & \begin{tabular}[c]{@{}c@{}}Prec\\ @200\end{tabular}  \\ \midrule
\checkmark & -           & -          & -           & 47.9       & 46.5      & 57.7   & 55.1    \\
\checkmark & \checkmark  & -          & -           & 53.4       & 54.3      & 61.9   & 60.3    \\
\checkmark & \checkmark  & \checkmark & -           & 54.3       & 54.9       & 62.9   & 61.3    \\
\checkmark & \checkmark  & -          & \checkmark  & 56.9       & 56.6       & 65.2   & 63.5   \\
\checkmark & \checkmark  & \checkmark & \checkmark  & \textbf{57.7}     & \textbf{57.7} & \textbf{65.8} &  \textbf{64.4}  \\ \bottomrule
\end{tabular}}
\end{table}

\subsection{Comparison with SOTA}
We compare the proposed method against existing state-of-the-art methods on both Sketchy Extended~\cite{yelamarthi2018zero} and TU-Berlin Extended~\cite{shen2018zero} datasets. We also report the experimental results of the methods that introduced the extra semantic embeddings. The quantitative results of different methods are reported in Table~\ref{tab:results}. Please note the backbone of the proposed method is ResNet-50. Our method outperforms existing state-of-the-art methods by a large margin even without any specially designed backbones~\cite{tian2021relationship} or semantic guidance~\cite{zhu2020ocean}. Our method has achieved 57.7\% $mAP@all$ and 65.8\% $Prec@100$ on TU-Berlin Extended dataset~\cite{yelamarthi2018zero}. The highest results of other methods are only 49.3\% $mAP@all$ and 61.2\% $Prec@100$, and our method has achieved 8.4\% improvement on $mAP@all$ and 4.6\% improvement on $Prec@100$. It is worth noting that our results of using the hashing codes even exceed previous highest performance. We observe that the proposed ACNet has achieved a larger performance gain on the TU-Berlin Extended dataset than the Sketchy Extended dataset. We attribute this phenomenon to the reason that the sketches from TU-Berlin Extended dataset have more abstract sketch representations and fewer fine-grained sketch details. Thus, the proposed ACNet can achieve a large performance gain by making the sketch domain approach to the photo domain. We provide more experimental results of using various embedding dimensions and backbone networks (VGG-16~\cite{simonyan2015very} and ResNet-50~\cite{he2016deep}) in appendix.

\begin{figure}
  \centering
  \includegraphics[width=\linewidth]{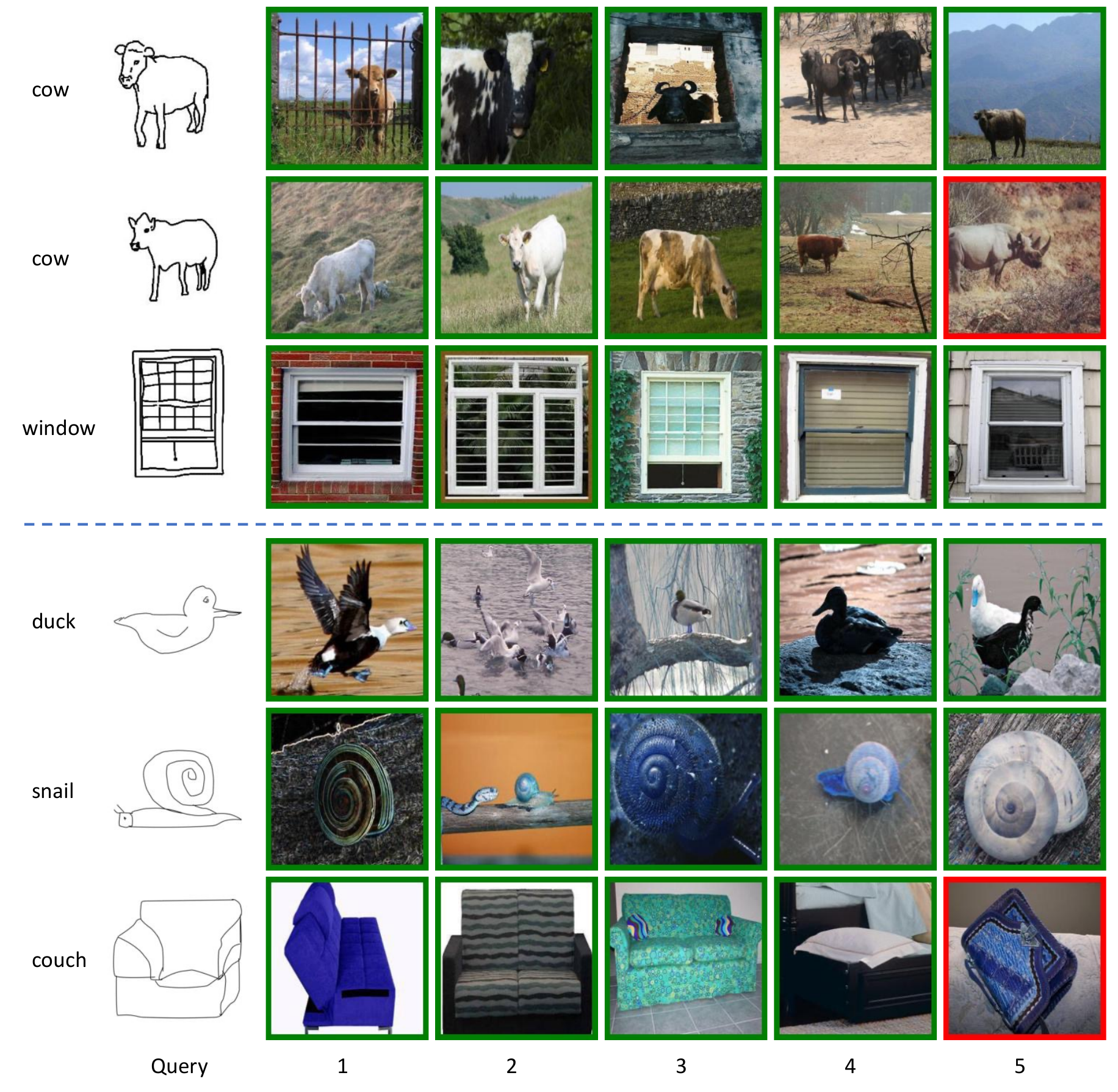}
  \caption{Top-5 ZS-SBIR retrieval results from the proposed model (ResNet-50 backbone with 512 embedding dimension) on the Sketchy Extended~\cite{yelamarthi2018zero} (first three rows) and TU-Berlin Extended~\cite{shen2018zero} (last three rows) datasets. Correct results are shown with a green border, while incorrect results are shown with a red border.}
  \label{fig:result}
\end{figure}

The qualitative results are shown in \YangWu{Figure~\ref{fig:result}}. We selected three instances from the two datasets to provide an intuitive comparison. The proposed \ywu{ACNet} could effectively find the correct photo \YangWu{given} a query sketch. We provide two similar instances from the same category ``cow'' and the two instances have the same orientation and similar shape representation except for the fine-grained representations on the head part. The proposed method could distinguish these fine-grained representations and provide corresponding desired photos rather than some very similar photos with prominent feature representations, which demonstrates that the proposed \ywu{ACNet} has indeed extracted some common knowledge among the unseen and seen categories. For a more challenging case: the ``couch'' sketch on the third row of TU-Berlin Extended dataset~\cite{shen2018zero}, the fifth retrieved photo belongs to ``purse'' even though the two categories are conceptually different. We attribute this failure to the reason that the retrieved wrong photo shares very similar structural representations with the real couch photos.

\subsection{Ablation Study}
\label{sec:ablation}
\noindent\textbf{Effectiveness of $\mathcal{L}_{adv}$ and $\mathcal{L}_{cha}$}.
Considering that $\mathcal{L}_{adv}$ and $\mathcal{L}_{cha}$ are inseparable, we aim to explore the effectiveness of the integration of the two loss functions. On the one hand, the joint training with $\mathcal{L}_{cha}$ could enable the gradients of the retrieval module to back-propagate to the synthesis module, which makes the synthesis better serve the retrieval module. On the other hand, the adversarial loss $\mathcal{L}_{adv}$ could help generate photo-like images with high data diversity and better align the sketch domain and the photo domain. We have achieved a huge retrieval performance improvement as shown in Table~\ref{tab:ablation_tuberlin} (from 46.5\% to 54.3\% on $mAP@all$) under the joint training manner with the GAN-based synthesis module. 

\noindent\textbf{Effectiveness of $\mathcal{L}_{ide}$}.
With the constraint of the pixel-level supervision from reconstructing the photo images, we have marginal improvement than the setting only with the $\mathcal{L}_{adv}$ and $\mathcal{L}_{cha}$. The $mAP@all$ score increases from  54.3\% to 54.9\% and the $Prec@100$ score increases by 1\% as shown in Table~\ref{tab:ablation_tuberlin}.

\noindent\textbf{Effectiveness of $\mathcal{L}_{norm}(\Phi(s_i))$}.
We have also removed the $\mathcal{L}_{norm}(\Phi(s_i))$ to explore the improvement of this component. Without this, there is a slight retrieval performance drop (from 57.7\% to 54.9\% on $mAP@all$) according to the quantitative scores reported in Table~\ref{tab:ablation_tuberlin}. More experimental \YangWu{analysis can} be found in appendix.

\subsection{Discussions}
\begin{figure}
  \centering
  \includegraphics[width=0.9\linewidth]{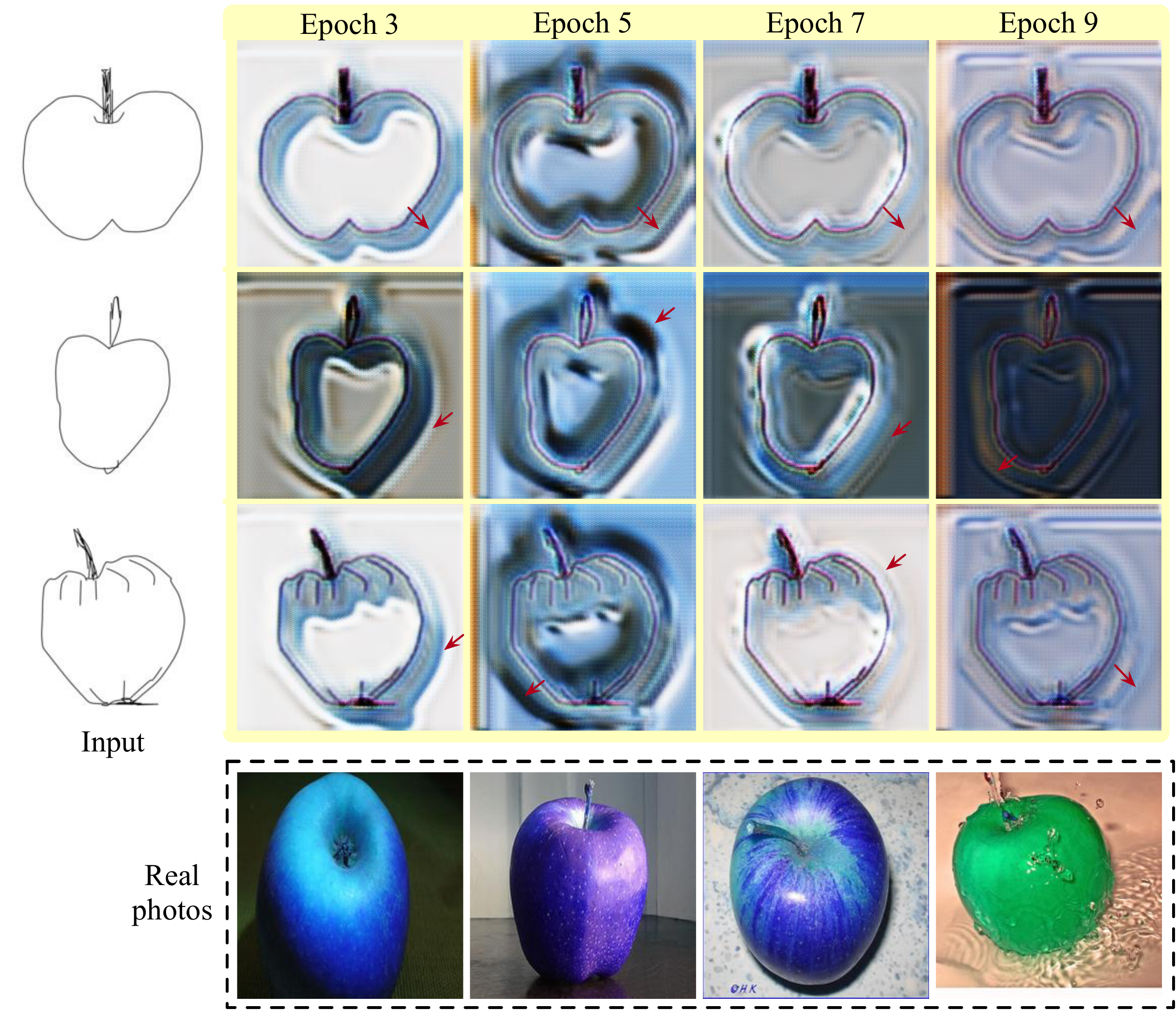}
  \caption{The qualitative sketch-to-photo synthesis results. Images in yellow box indicate the intermediate synthesized images. We also provide some real photos for reference. The sketch-to-photo generator could generate similar shape representations with the real apple photos. Best viewed in color.}
  \label{fig:syns}
\end{figure}

\noindent\textbf{Qualitative Sketch-to-photo Synthesis}.
We provide some visual synthesis results of our sketch-to-photo generator at different epochs in \YangWu{Figure}~\ref{fig:syns}. As illustrated, the synthesized images still have visual artifacts, which are not \textit{human-friendly}. We observe the generator has generated obvious boundary (pointed by red arrows) representations, which match the real apple shape better. Thus, we guess that with the constraint of the chainer loss, $G$ aims to synthesize more prominent feature representations, which are more \textit{algorithm-friendly}, recognizable and comparable for $\Phi$. Finally, we also argue that our goal is to promote retrieval performance rather than improve the naturalness and aesthetic quality of the synthesized samples.

\begin{figure}
  \centering
  \includegraphics[width=0.9\linewidth]{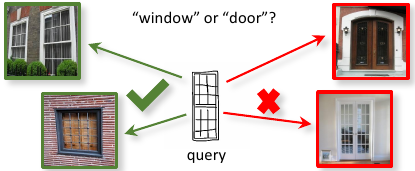}
  \caption{Ambiguous case: the given sketch has the similar structure as ``window'' and ``door''.}
  \label{fig:case}
\end{figure}

\noindent\textbf{Limitations}.
When given a sketch query that is highly succinct and abstract in \YangWu{Figure}~\ref{fig:case}, it is extremely challenging to judge whether it is ``window'' or ``door'' due to the ambiguous correspondences to the photos. Since the sketch-to-photo synthesis module of our method may still fail to generate desired reasonable images with high confidence, our method cannot handle this case well, either.

\noindent\textbf{Broader Impact}.
Our method has pioneered a new direction to combine synthesis and the downstream visual retrieval task through a joint training manner, in which the gradients of the downstream vision task could be propagated to the synthesis module to tell the generator how to synthesize. Our joint training framework has the potential to be extended to other downstream vision tasks such as semantic segmentation, object detection under some challenging environments (\eg, nighttime, foggy and rainy night), in which the collected visual images have poor visibility. Similarly, we could perform the image synthesis to enhance the image quality to better serve the downstream vision tasks.

\section{Conclusion}
In this work, we proposed a novel, simple and effective joint synthesis-and-retrieval network \ywu{called Approaching-and-Centralizing Network (ACNet)} for ZS-SBIR and achieved state-of-the-art performance on Sketchy Extended~\cite{yelamarthi2018zero} and TU-Berlin Extended~\cite{shen2018zero} datasets. The joint training of the sketch-to-photo synthesis module and the retrieval module could effectively reduce both the domain gap and knowledge gap. The chainer loss is designed to force the retrieval 
module to extract more reliable and effective feature representations from the synthesized samples even with noise and visual artifacts. The generalization ability to unseen categories can also be enhanced by constantly generating samples with high diversity. Finally, by propagating the gradients of the retrieval module to the synthesis module, we can also tell the generator how to generate samples, which can better serve the downstream vision task. Our joint training framework provides valuable insight on how to integrate the synthesis and other vision tasks.

{\small
\bibliographystyle{ieee_fullname}
\bibliography{egbib}
}

\section*{Appendix}
\appendix

In \YangWu{this} appendix, we first provide the implementation details of our sketch-to-photo generator and corresponding discriminator. Then we provide more experimental results to demonstrate that the NormSoftmax is better than Triplet and the proposed joint training \YangWu{scheme} is more effective than two-stage training \YangWu{scheme} for the ZS-SBIR task. The results of using various backbones and embedding sizes are also included. More qualitative results are provided to give an intuitive comparison. Finally, we provide \YangWu{further} ablation studies to explore the sensitiveness and effectiveness of the proposed ACNet to different hyper parameters.

\section{Implementation Details}
We follow the generator architecture of the vanilla CycleGAN~\cite{zhu2017unpaired} and design our sketch-to-photo generator. 

\textbf{Generator architecture}. For $G$, we first adopt the Conv-InstanceNorm-ReLU (\textbf{CIR} for short) block with kernel size 7 and stride size 1 for the sketches. Then two CIR blocks with kernel size 3 and stride size 2 to achieve down-sampling are conducted. To enlarge the information capacity, $G$ combines 8 residual blocks (\textbf{RB}) to stack residual information. For the residual blocks, the kernel size is 3 and the stride size is 1. As for the reverse up-sampling stage, another 3 Deconv-InstanceNorm-ReLU (\textbf{DIR}) blocks are adopted. Finally, a Tanh activation is adopted to obtain the normalized photo-like images. 

\textbf{Discriminator architecture}. We first apply one Conv-LeakyReLU layer (\textbf{CLR}) with kernel size 4 and stride size 2 to achieve down-sampling. Then three Conv-InstanceNorm-LeakyReLU layers (\textbf{CILR}) with kernel size 4 are adopted. The slope for the LeakyReLU is 0.2. We finally obtain the logit output to perform the real/fake discrimination after one convolution layer with kernel size 4.

We refer the readers to check Table~\ref{tab:generator} and Table~\ref{tab:discriminator} for more details. We also provide the corresponding ZS-SBIR experiments of using different architectures of $G$ in Section~\ref{sec:cap}.

\begin{table}
\centering
\caption{The network architecture of $G$. $c$ is the \YangWu{number of channels} of the first convolution layer.}
\label{tab:generator}
\small
\setlength{\tabcolsep}{8pt}
\begin{tabular}{c|cccc}
\toprule
\multicolumn{5}{c}{$G$ (sketch-to-photo generator)}  \\\midrule 
\begin{tabular}[c]{@{}c@{}}Module\\ Type\end{tabular}      & \begin{tabular}[c]{@{}c@{}}Kernel\\ Size\end{tabular} & \begin{tabular}[c]{@{}c@{}}Stride\\ Size\end{tabular} & \begin{tabular}[c]{@{}c@{}}Channel\\ Number\end{tabular} & \begin{tabular}[c]{@{}c@{}}Padding\\ Type\end{tabular} \\
\midrule
\multirow{3}{*}{CIR}     & 7           & 1      & $c=8$  & reflect \\ 
     & 3           & 2      & 16 ($2\times c$) & zero \\ 
     & 3           & 2      & 32 ($4\times c$) & zero \\ \midrule
\multirow{8}{*}{RB} & 3           & 1      & 32 ($4\times c$)  & reflect \\ 
& 3           & 1      & 32 ($4\times c$)  & reflect \\ 
& 3           & 1      & 32 ($4\times c$) & reflect \\ 
& 3           & 1      & 32 ($4\times c$) & reflect \\ 
& 3           & 1      & 32 ($4\times c$) & reflect \\ 
& 3           & 1      & 32 ($4\times c$) & reflect \\ 
& 3           & 1      & 32 ($4\times c$) & reflect \\ 
& 3           & 1      & 32 ($4\times c$) & reflect \\ \midrule
\multirow{3}{*}{DIR}   & 3           & 2      & 16 ($2\times c$)  & zero  \\
& 3           & 2      & 8      & zero \\ 
& 7           & 1      & 3 & reflect \\ \midrule
Tanh   & -           & -      & 3 & reflect \\ \bottomrule
\end{tabular}
\end{table}

\begin{table}
\centering
\caption{The network architecture of $D$. $c$ is the \YangWu{number of channels} of the first convolution layer.}
\label{tab:discriminator}
\small
\setlength{\tabcolsep}{9pt}
\begin{tabular}{c|ccc}
\toprule
\multicolumn{4}{c}{$D$ (Discriminator)}  \\ \midrule 
\begin{tabular}[c]{@{}c@{}}Module\\ Type\end{tabular}  & \begin{tabular}[c]{@{}c@{}}Kernel\\ Size\end{tabular} & \begin{tabular}[c]{@{}c@{}}Stride\\ Size\end{tabular}  & \begin{tabular}[c]{@{}c@{}}Channel\\ Number\end{tabular} \\
\midrule
CLR & 4  & 2      & $c=8$ \\ \midrule
\multirow{3}{*}{CILR} & 4  & 2      & 16 ($2\times c$)\\
& 4  & 2      & 32 ($4\times c$) \\ 
& 4  & 1      & 64 ($8\times c$) \\ \midrule
Conv & 4  & 1      & 1  \\ \bottomrule
\end{tabular}
\end{table}

\section{More Experiments}
\subsection{Triplet vs. NormSoftmax}
Following the same experimental setting on the Sketchy Extended~\cite{yelamarthi2018zero} dataset \YangWu{as presented in the main text}, we chose the vanilla Triplet loss and NormSoftmax loss and perform the ZS-SBIR experiments on the TU-Berlin Extended~\cite{shen2018zero} dataset. The quantitative results are reported in Table~\ref{tab:prior_tuberlin}. We can observe that the NormSoftmax outperforms the Triplet by a large margin (47.9\% vs. 38.1\% on $mAP@200$ score), which indicates that the proxy-based loss has a significant priority over the triplets-based loss on ZS-SBIR task.

\begin{figure*}
  \centering
  \includegraphics[width=\linewidth]{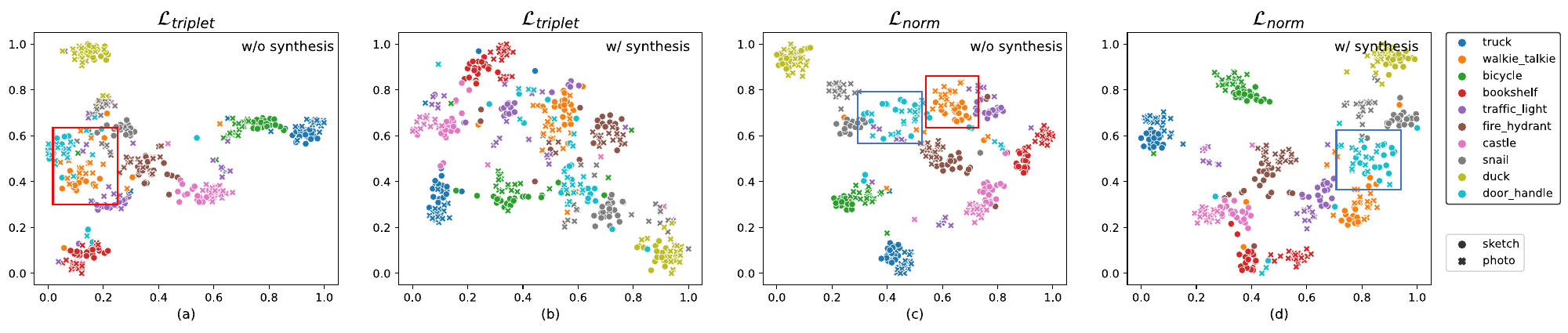}
  \caption{T-SNE visualization of sketch and photo embeddings on TU-Berlin Extended dataset. We randomly choose 20 samples from 10 unseen test categories for visualization. Different colors refer to different categories. The two-stage training strategy cannot obtain more separable embeddings than the vanilla setting, regardless of using Triplet loss or NormSoftmax loss. We refer the readers to pay more attention to the regions covered by the same color boxes for better comparison.}
  \label{fig:tsne_tuberlin}
\end{figure*}

\subsection{Two-stage Training vs. Joint Training}
For the two-stage training strategy, we first adopt the vanilla CycleGAN~\cite{zhu2017unpaired} to perform the unpaired image-to-image (I2I) translation between the sketches and photos on the TU-Berlin Extended dataset. After the unpaired I2I model converges, we adopt the trained sketch-to-photo generator for inference and translate all the sketches into photo-like images. Please note, both training and test sketches have been translated to the photo domain for further training and testing. We then perform the ZS-SBIR experiments based on the synthesized photo-like images and the real photos. The original category labels of the sketches are inherited for training. We choose both Triplet loss and the adopted NormSoftmax loss for optimization and the quantitative results are also reported in Table~\ref{tab:prior_tuberlin}. 

We can obtain better results based on $\mathcal{L}_{norm}$ by approaching the sketch domain to the photo domain. We attribute this improvement to the reason that the sketch-to-photo synthesis module could help the retrieval module capture more effective feature representations. In contrast, there is a performance drop while using the Triplet loss for optimization, which also indicates that the NormSoftmax~\cite{zhai2019classification} has a stronger ability to coordinate the real samples and the synthesized samples than Triplet~\cite{ge2018deep}. The NormSoftmax is more compatible with the sketch-to-photo synthesis for data transformation to mitigate the domain gap.

\begin{table}
\centering
\caption{The ZS-SBIR performance comparison under various settings on TU-Berlin Extended dataset: 1) Triplet and NormSoftmax for optimization; and 2) with or without sketch-to-photo synthesis through CycleGAN to mitigate the domain gap.}
\label{tab:prior_tuberlin}
\small
\setlength{\tabcolsep}{7.5pt}
\begin{tabular}{c|c|cccc}
\toprule
\begin{tabular}[c]{@{}c@{}}Loss\\ Type\end{tabular} & Synthetic & \begin{tabular}[c]{@{}c@{}}mAP\\ @200\end{tabular} & \begin{tabular}[c]{@{}c@{}}mAP\\ @all\end{tabular} & \begin{tabular}[c]{@{}c@{}}Prec\\ @100\end{tabular} & \begin{tabular}[c]{@{}c@{}}Prec\\ @200\end{tabular} \\ \midrule
\multirow{2}{*}{$\mathcal{L}_{triplet}$} & -   & \textbf{38.1}     & \textbf{36.8}       & \textbf{49.8}       & \textbf{47.1}         \\
 & \checkmark   & 36.3     & 35.2       & 47.7       & 45.4         \\\midrule
\multirow{2}{*}{$\mathcal{L}_{norm}$} & -   & 47.9 & 46.5       & 57.7       & 55.1         \\
 & \checkmark & \textbf{48.4}     & \textbf{46.6}       & \textbf{58.0}     & \textbf{55.6}         \\ \bottomrule
\end{tabular}
\end{table}

\begin{figure}
  \centering
  \includegraphics[width=0.9\linewidth]{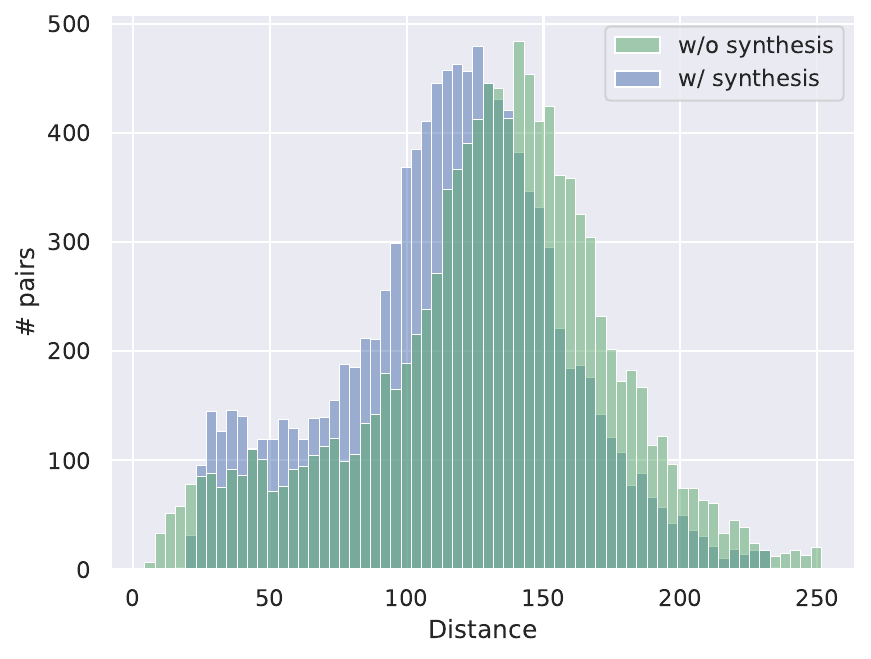}
  \caption{Histograms of sample distances under two settings: 1) between original sketches and real photos (``w/o synthesis'') and 2) between synthesized images and real photos (``w/ synthesis''). The experiments are conducted on the TU-Berlin Extended dataset. Since the sketches of this dataset are highly abstract, succinct and exaggerated, the vanilla CycleGAN could only mitigate the domain gap to some extent.}
  \label{fig:prior_tuberlin}
\end{figure}

We further compute the $L_1$ distances between 50 randomly sampled instances from each training category under two settings: 1) between the original sketches and the real photos and 2) between the synthesized images and the real photos. We provide the distance histogram in Figure~\ref{fig:prior_tuberlin} to show the domain distance. Since the sketches from the TU-Berlin Extended dataset are highly abstract, succinct and exaggerated, with the synthesis, we can reduce the domain gap to some extent. 

\begin{figure}
  \centering
  \includegraphics[width=\linewidth]{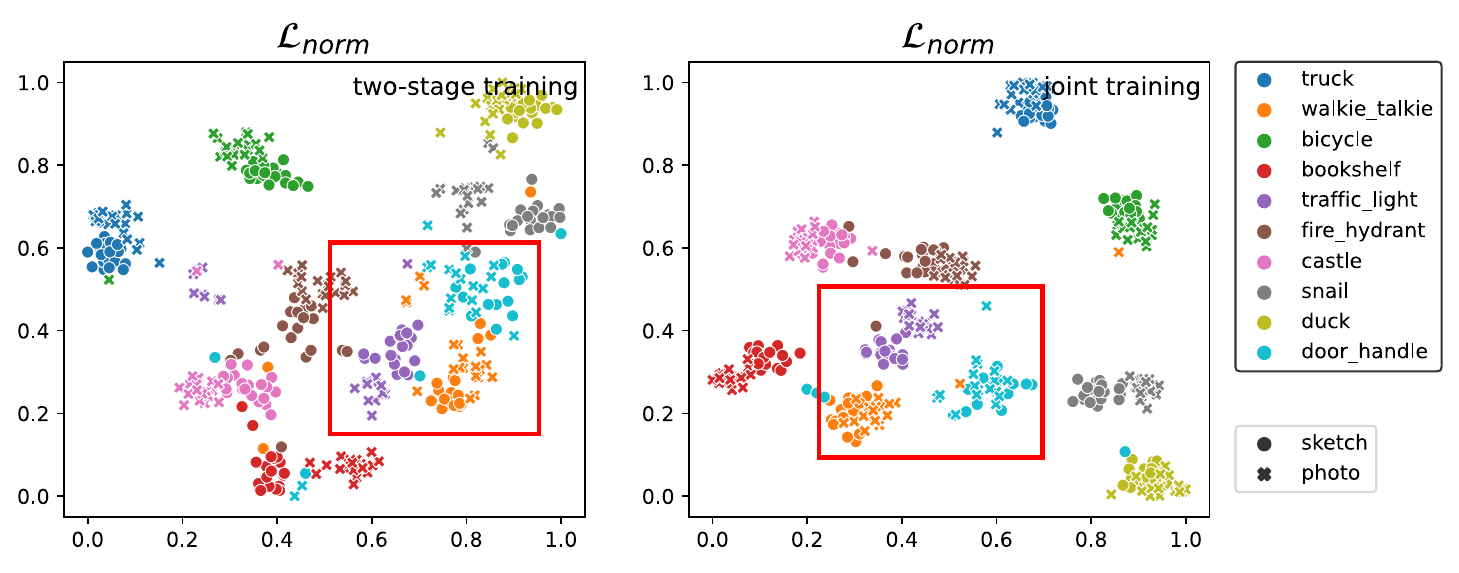}
  \caption{T-SNE visualization of sketch and photo embeddings on TU-Berlin Extended dataset under the two-stage training and joint training. We refer the readers to pay more attention on the regions covered by same color boxes for better comparison.}
  \label{fig:tuberlin_ours}
\end{figure}

Furthermore, we provide the T-SNE visualization of sketch and photo embeddings on TU-Berlin Extended dataset in Figure~\ref{fig:tsne_tuberlin} under the four settings of Table~\ref{tab:prior_tuberlin}. The two-stage training strategy cannot obtain much more separable embeddings to achieve a large ZS-SBIR performance gain. We attribute this failure to the reason that the gradients of the retrieval module cannot be directly been utilized for the optimization of the synthesis module. Thus, the synthesis module has no idea how to generate images, which can better serve the image retrieval. We provide a direct comparison between the distribution of the embeddings from the two-stage training and the proposed joint training in Figure~\ref{fig:tuberlin_ours}. As illustrated, our joint training has a strong ability to split the embeddings from different classes, which leads to better ZS-SBIR performance.

\subsection{Various Backbones and Embedding Sizes}
In this section, we aim to explore the effectiveness and sensitiveness of choosing various backbone networks: ResNet-50~\cite{he2016deep} and VGG-16~\cite{simonyan2015very}. We conduct the ZS-SBIR experiments on both the Sketchy Extended and TU-Berlin Extended datasets. All the quantitative results under various settings are reported in Table~\ref{tab:ablation_backbone}. With the same embedding size, we can obtain better results based on the ResNet-50 than the VGG-16. The proposed ACNet could achieve a very impressive 58.6\% $mAP@all$ score on the TU-Berlin Extended dataset by choosing the ResNet-50 with 4096 embedding size as the backbone network.

\begin{table}
\centering
\caption{Overall comparison of our method with different backbone networks and embedding sizes on Sketchy Extended and TU-Berlin Extended datasets. The best scores are bolded.}
\label{tab:ablation_backbone}
\small
\setlength{\tabcolsep}{5pt}
\begin{tabular}{c|c|cc|cc}
\toprule
\multirow{3}{*}{Backbone} & \multirow{3}{*}{Dim} & \multicolumn{2}{c|}{Sketchy Extended} & \multicolumn{2}{c}{TU-Berlin Extended} \\
&     & \begin{tabular}[c]{@{}c@{}}mAP\\ @200\end{tabular}  & \begin{tabular}[c]{@{}c@{}}Prec\\ @200\end{tabular} & \begin{tabular}[c]{@{}c@{}}mAP\\ @all\end{tabular} & \begin{tabular}[c]{@{}c@{}}Prec\\ @100\end{tabular} \\ \midrule
\multirow{3}{*}{VGG-16}    & 64      &  32.6    &  44.7       &  37.1    &  50.6  \\ 
& 512      &   38.3    &      49.3   &   43.9   & 58.1  \\
& 4096      &   \textbf{40.0}    &      \textbf{50.8}   &   \textbf{47.9}   & \textbf{62.3}    \\  \midrule
\multirow{3}{*}{ResNet-50}    & 64      &  43.0    &      52.7   &   44.9   & 57.2  \\ 
& 512      &   \textbf{51.7}    &      \textbf{60.8}   &   57.7   & \textbf{65.8}  \\
& 4096      &   51.1    &      60.0   &   \textbf{58.6}   & 64.6    \\ \bottomrule
\end{tabular}
\end{table}

\subsection{More Qualitative Results}
\begin{figure*}
  \centering
  \includegraphics[width=\linewidth]{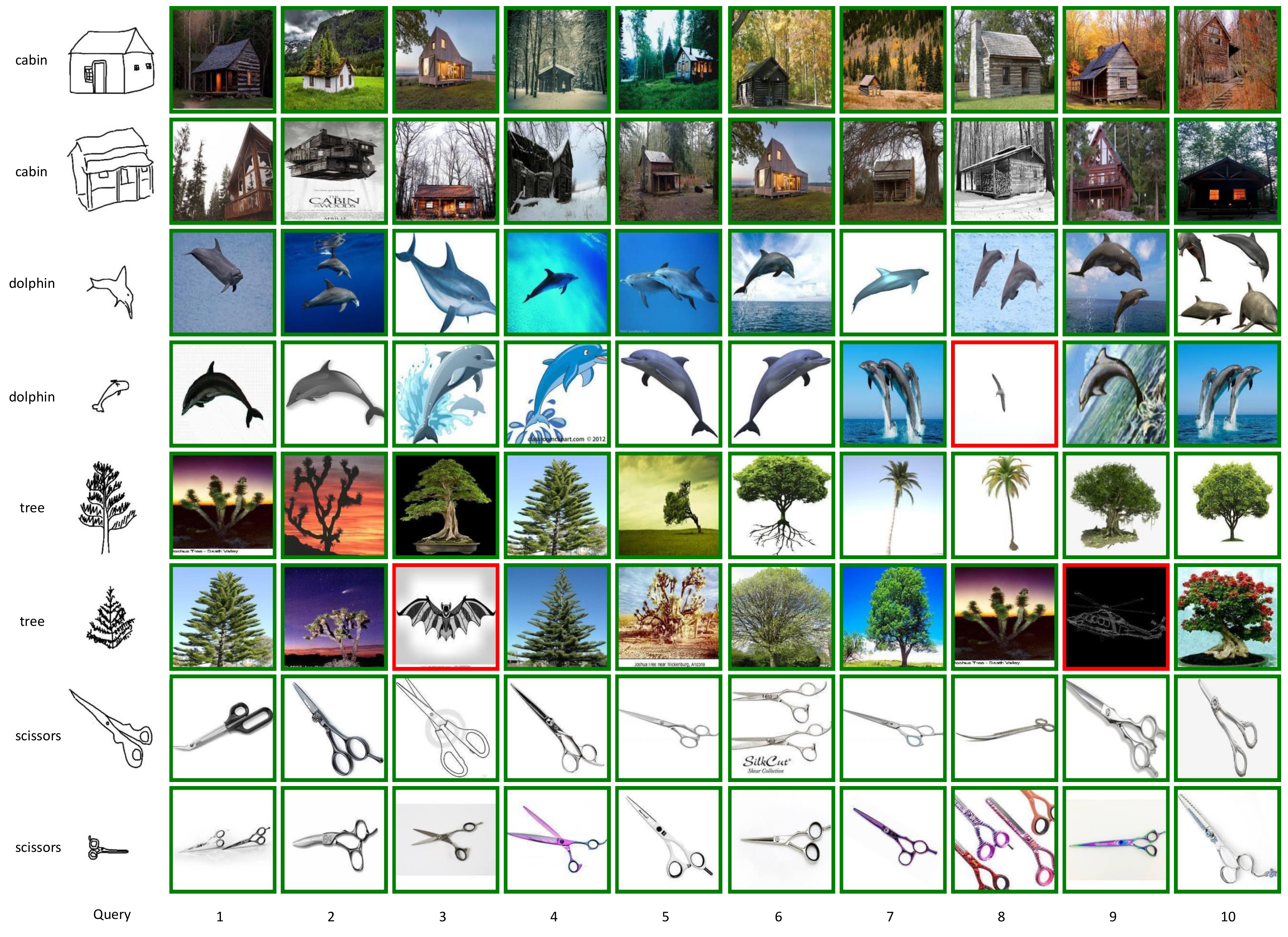}
  \caption{Top-10 ZS-SBIR retrieval results from the proposed model (ResNet-50 backbone with 512 embedding dimension) on the Sketchy Extended~\cite{yelamarthi2018zero} dataset. Correct results are shown with a green border, while incorrect results are shown with a red border.}
  \label{fig:result_sketchy}
\end{figure*}

\begin{figure*}
  \centering
  \includegraphics[width=\linewidth]{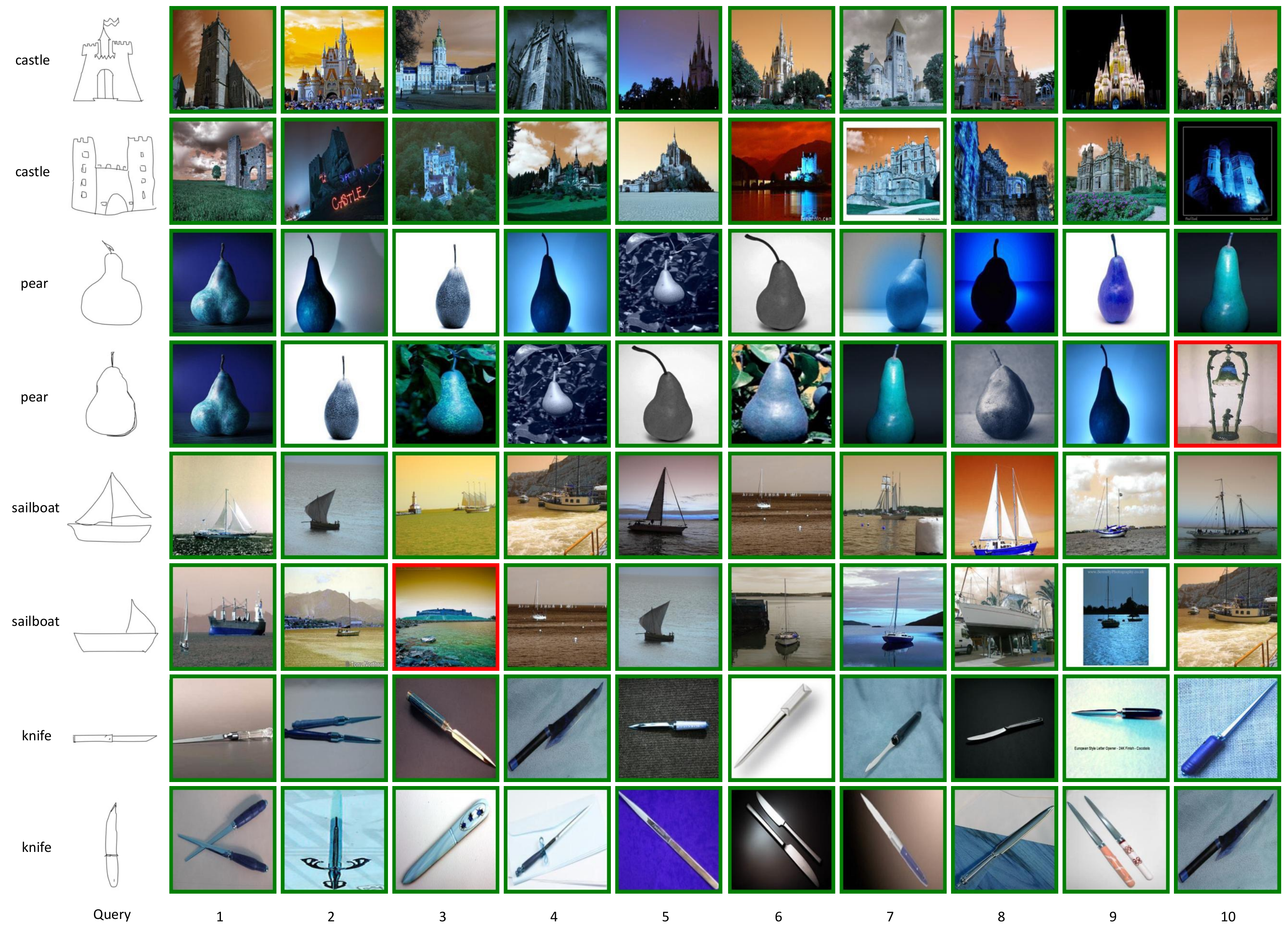}
  \caption{Top-10 ZS-SBIR retrieval results from the proposed model (ResNet-50 backbone with 512 embedding dimension) on the TU-Berlin Extended~\cite{shen2018zero} dataset. Correct results are shown with a green border, while incorrect results are shown with a red border.}
  \label{fig:result_tuberlin}
\end{figure*}

We provide more visual retrieval results on the Sketchy Extended and TU-Berlin Extended datasets in Figure~\ref{fig:result_sketchy} and Figure~\ref{fig:result_tuberlin}, respectively. We selected eight instances from the two datasets to provide an intuitive comparison. The proposed ACNet could effectively find the correct photo given a query sketch. 

\section{Ablation Studies}
\subsection{Component Dissection}
We have also performed the ablation studies on the Sketchy Extended dataset to show the effectiveness of each component of the proposed ACNet. All the quantitative results are reported in Table~\ref{tab:ablation_sketchy}. With only the NormSoftmax loss, the proposed method could only achieve a 45.2\% $mAP@200$ score. By combining the sketch-to-photo synthesis in a joint training manner, we can increase the $mAP@200$ score from 45.2\% to 47.8\%. The identity loss could also help promote the synthesis quality and thus lead to a performance gain (from 47.8\% to 48.4\% on $mAP@200$ score). Finally, with the constraint of $\mathcal{L}_{norm}(\Phi(s_i))$, the proposed method could better mitigate the domain gap between the sketches and the photos.

\begin{table}
\setlength\tabcolsep{3pt}
\centering
\caption{Ablation studies for the proposed method on Sketchy Extended dataset. We adopt the ResNet-50 as our backbone and the embedding dimension is 512. The NormSoftmax loss is used as a vanilla baseline.}
\label{tab:ablation_sketchy}
\small
\setlength{\tabcolsep}{0.5pt}
\begin{tabular}{cccc|cccc}
\toprule
{\footnotesize $\mathcal{L}_{norm}$ } & {\footnotesize $\mathcal{L}_{adv}\&\mathcal{L}_{cha}$ }  & {\footnotesize $\mathcal{L}_{ide}$ } & {\footnotesize $\mathcal{L}_{norm}(\Phi(s_i))$ } & \begin{tabular}[c]{@{}c@{}}mAP\\ @200\end{tabular}   &  \begin{tabular}[c]{@{}c@{}}mAP\\ @all\end{tabular}   & \begin{tabular}[c]{@{}c@{}}Prec\\ @100\end{tabular}  & \begin{tabular}[c]{@{}c@{}}Prec\\ @200\end{tabular}  \\ \midrule
\checkmark & -           & -          & -           & 45.2       & 48.6       & 60.2   & 55.7    \\
\checkmark & \checkmark  & -          & -           & 47.8       & 52.7       &  60.9  & 57.1    \\
\checkmark & \checkmark  & \checkmark & -           & 48.4       & 53.0       &  61.4  & 57.6   \\
\checkmark & \checkmark  & -          & \checkmark  & 49.7       & 53.7       &  62.2  & 58.7   \\
\checkmark & \checkmark  & \checkmark & \checkmark  & \textbf{51.7}       & \textbf{55.9}       &  \textbf{64.3}  & \textbf{60.8}   \\ \bottomrule
\end{tabular}
\end{table}

\subsection{GAN Capacity}
\label{sec:cap}
We also explore the influences of choosing different architectures for $G$ and $D$ on the final ZS-SBIR results. We conduct the corresponding experiments in two ways. We first set the \YangWu{number of channels} $c$ to different values for both $G$ and $D$. The quantitative results are reported in Table~\ref{tab:ablation_channel}. We observe that we can achieve better results when using a lightweight architecture for $G$ and $D$. We guess that it is more possible for the model with a bigger network capacity to introduce the noise and uncertainty for the downstream retrieval module.

Later, we design different architectures for $G$ by choosing different numbers of residual blocks and the results are reported in Table~\ref{tab:ablation_block}. An appropriate number (\eg, 6 or 8) of residual block could achieve the best ZS-SBIR results.

\begin{table}
\centering
\caption{Overall comparison of our method with different \YangWu{number of channels} on Sketchy Extended and TU-Berlin Extended datasets. The best and second-best results are bold and underlined, respectively.}
\label{tab:ablation_channel}
\small
\setlength{\tabcolsep}{8.5pt}
\begin{tabular}{c|cc|cc}
\toprule
\multirow{3}{*}{\begin{tabular}[c]{@{}c@{}}\YangWu{Number of}\\ \YangWu{Channels} ($c$)\end{tabular}} & \multicolumn{2}{c|}{Sketchy Extended} & \multicolumn{2}{c}{TU-Berlin Extended} \\
& \begin{tabular}[c]{@{}c@{}}mAP\\ @200\end{tabular}  & \begin{tabular}[c]{@{}c@{}}Prec\\ @200\end{tabular} & \begin{tabular}[c]{@{}c@{}}mAP\\ @all\end{tabular} & \begin{tabular}[c]{@{}c@{}}Prec\\ @100\end{tabular} \\ \midrule
4     & \textbf{51.9}    &  \textbf{60.9}   &  56.8   & 65.5  \\ 
8    & 51.7    &  \underline{60.8}   &  \textbf{57.7}   & \underline{65.8}  \\
16    & \underline{51.8}    &  60.5   &  \underline{56.9}   & \textbf{66.1}   \\ 
32    & 50.9    &  60.1   &  56.3   & 65.1  \\
64    & 49.0    &  58.5   &  54.1   & 62.6  \\ \bottomrule
\end{tabular}
\end{table}

\begin{table}
\centering
\caption{Overall comparison of our method with different number of blocks on Sketchy Extended and TU-Berlin Extended datasets. The best and second-best results are bold and underlined, respectively.}
\label{tab:ablation_block}
\small
\setlength{\tabcolsep}{9.5pt}
\begin{tabular}{c|cc|cc}
\toprule
\multirow{3}{*}{\begin{tabular}[c]{@{}c@{}}\YangWu{Number of} \\ \YangWu{Blocks} \end{tabular}} & \multicolumn{2}{c|}{Sketchy Extended} & \multicolumn{2}{c}{TU-Berlin Extended} \\
& \begin{tabular}[c]{@{}c@{}}mAP\\ @200\end{tabular}  & \begin{tabular}[c]{@{}c@{}}Prec\\ @200\end{tabular} & \begin{tabular}[c]{@{}c@{}}mAP\\ @all\end{tabular} & \begin{tabular}[c]{@{}c@{}}Prec\\ @100\end{tabular} \\ \midrule
4    & \underline{50.5}    &  \underline{59.8}   &  55.9   & \underline{65.1}  \\
6    & 50.1    &  59.2   &  \textbf{57.8}   & \textbf{65.8}   \\ 
8    & \textbf{51.7}    &  \textbf{60.8}   &  \underline{57.7}   & \textbf{65.8}  \\
9    & 49.6    &  59.1   &  55.4   & 64.6  \\ \bottomrule
\end{tabular}
\end{table}

\begin{table}
\centering
\caption{Overall comparison of our method with different hyper parameters on Sketchy Extended dataset and TU-Berlin Extended dataset.}
\label{tab:ablation_hyper}
\small
\setlength{\tabcolsep}{9pt}
\begin{tabular}{c|c|cc|cc}
\toprule
\multirow{3}{*}{$\lambda$} &\multirow{3}{*}{$\gamma$} & \multicolumn{2}{c|}{Sketchy Extended} & \multicolumn{2}{c}{TU-Berlin Extended} \\
&     & \begin{tabular}[c]{@{}c@{}}mAP\\ @200\end{tabular}  & \begin{tabular}[c]{@{}c@{}}Prec\\ @200\end{tabular} & \begin{tabular}[c]{@{}c@{}}mAP\\ @all\end{tabular} & \begin{tabular}[c]{@{}c@{}}Prec\\ @100\end{tabular} \\ \midrule
0.1   & 10     & 46.9    &  56.7   &  54.9   & 63.7  \\ 
1.0   & 1.0    & 49.9    &  59.2   &  56.5   & 65.2  \\
10    & 0.1    & \textbf{51.7}    &  \textbf{60.8}   &  \textbf{57.7}   & \textbf{65.8}   \\ \bottomrule
\end{tabular}
\end{table}

\subsection{Selection of $\lambda$ and $\gamma$}
Rethink our final objective function:
\begin{equation}
    \mathcal{L}(G,D,\Phi)=\mathcal{L}_{adv}(G,D)+ \lambda\mathcal{L}_{norm}+\gamma\mathcal{L}_{ide},
    \label{eq:all}
\end{equation}
there are two hyper parameters to balance the contribution of each loss function. To explore the sensitiveness of the proposed ACNet to the two hyper parameters, we have designed three experiments of using different combinations of $\lambda$ and $\gamma$: $(\lambda=0.1,\gamma=10)$, $(\lambda=1.0,\gamma=1.0)$ and $(\lambda=10,\gamma=0.1)$. The quantitative comparison is reported in Table~\ref{tab:ablation_hyper}. As observed, the proposed ACNet could achieve the highest scores when $\lambda=10$ and $\gamma=0.1$.

\end{document}